\documentclass[letterpaper]{article}
\usepackage{aaai}
\usepackage{times}
\usepackage{helvet}
\usepackage{courier}
\usepackage{graphicx}

\newcommand{\eqref}[1]{Eq.~(\ref{#1})}
\newcommand{\figref}[1]{Figure~\ref{#1}}
\newcommand{\tabref}[1]{Table~\ref{#1}}

\newcommand{\commentout}[1]{%
}

\begin{document}

\title{Exploiting Social Annotation for Automatic Resource Discovery
}
\author{Anon Plangprasopchok and Kristina Lerman\\
USC Information Sciences Institute\\
4676 Admiralty Way \\
Marina del Rey, CA 90292, USA \\
\{plangpra,lerman\}@isi.edu}

\maketitle

\begin{abstract}
Information integration applications, such as mediators or mashups,
that require access to information resources currently rely on users
manually discovering and integrating them in the application. Manual
resource discovery is a slow process, requiring the user to sift
through results obtained via keyword-based search. Although search
methods have advanced to include evidence from document contents,
its metadata and the contents and link structure of the referring
pages, they still do not adequately cover information sources ---
often called ``the hidden Web''--- that dynamically generate
documents in response to a query. The recently popular social
bookmarking sites, which allow users to annotate and share metadata
about various information sources, provide rich evidence for
resource discovery. In this paper, we describe a probabilistic model
of the user annotation process in a social bookmarking system
\emph{del.icio.us}. We then use the model to automatically find
resources relevant to a particular information domain. Our
experimental results on data obtained from \emph{del.icio.us} show
this approach as a promising method for helping automate the
resource discovery task.

\end{abstract}

\section{Introduction}

As the Web matures, an increasing number of dynamic information
sources and services come online. Unlike static Web pages, these
resources generate their contents dynamically in response to a
query. They can be HTML-based, searching the site via an HTML
form, or be a Web service. Proliferation of such resources has led
to a number of novel applications, including Web-based mashups,
such as Google maps and Yahoo pipes, information integration
applications~\cite{Thakkar2005} and intelligent office assistants
~\cite{Lerman07ijswis} that compose information resources within
the tasks they perform. In all these applications, however, the
user must discover and model the relevant resources. Manual
resource discovery is a very time consuming and laborious process.
The user usually queries a Web search engine with appropriate
keywords and additional parameters (e.g., asking for .kml or .wsdl
files), and then must examine every resource returned by the
search engine to evaluate whether it has the desired
functionality. Often, it is desirable to have not one but several
resources with an equivalent functionality to ensure robustness of
information integration applications in the face of resource
failure. Identifying several equivalent resources makes manual
resource discovery even more time consuming.

The majority of the research in this area of information
integration has focused on automating modeling resources --- i.e.,
understanding semantics of data  they
use~\cite{Hess2003,Lerman2006aaai} and the functionality they
provide~\cite{Carman2007}. In comparison, the resource discovery
problem has received much less attention. Note that traditional
search engines, which index resources by their contents --- the
words or terms they contain
--- are not likely to be useful in this
domain, since the contents is dynamically generated. At best, they
rely on the metadata supplied by the resource authors or the
anchor text in the pages that link to the resource.
Woogle~\cite{Woogle} is one of the few search engines to index Web
services based on the syntactic metadata provided in the WSDL
files. It allows a user to search for services with a similar
functionality or that accept the same inputs as another services.

Recently, a new generation of Web sites has rapidly gained
popularity. Dubbed ``social media,'' these sites allow users to
share documents, including bookmarks, photos, or videos, and to
\emph{tag} the content with free-form keywords. While the initial
purpose of tagging was to help users organize and manage their own
documents, it has since been proposed that collective tagging of
common documents can be used to organize information via an
informal classification system dubbed a
``folksonomy''~\cite{folksonomy}. Consider, for example,
\emph{http://geocoder.us}, a geocoding service that takes an input
as address and returns its latitude and longitude. On the social
bookmarking site \emph{del.icio.us}\footnote{http://del.icio.us},
this resource has been tagged by more than $1,000$ people. The
most common tags associated by users with this resource are
``map,'' ``geocoding,'' ``gps,'' ``address,'' ``latitude,'' and
``longitude.'' This example suggests that although there is
generally no controlled vocabulary in a social annotation system,
tags can be used to categorize resources by their functionality.

We claim that social tagging can be used for information resource
discovery. We explore three probabilistic generative models that
can be used to describe the tagging process on \emph{del.icio.us}.
The first model is the probabilistic Latent Semantic
model~\cite{HofmannPLSA99} which ignores individual user by
integrating bookmarking behaviors from all users.
%The other is the three-ways aspect model
%, which was originally applied in
%recommendation system \cite{Popescul01}.
The second model, the three-way aspect model, was
proposed~\cite{ChineseModel} to model \emph{del.icio.us} users'
annotations. The model assumes that there exists a global
conceptual space that generates the observed values for users,
resources and tags independently. We propose an alternative third
model, motivated by the Author-Topic model
\cite{TopicModelSmyth2004}, which maintains that latent topics
that are of interest to the author generate the words in the
documents. Since a single resource on \emph{del.icio.us} could be
tagged differently by different users, we separate ``topics'', as
defined in Author-Topic model, into ``(user) interests'' and
``(resource) topics''. Together user interests and resource topics
generate tags for one resource. In order to use the models for
resource discovery, we describe each resource by a topic
distribution and then compare this distribution with that of all
other resources in order to identify relevant resources.

The paper is organized as follows. In the next section, we
describe how tagging data is used in resource discovery. Subsequently
we present the probabilistic model we have developed to
aid in the resource discovery task. The section also describes two
earlier related models. We then compares the performance of the
three models on the datasets obtained from \emph{del.icio.us}. We
review prior work and finally present conclusions and future research
directions.

\section{Problem Definition}
\label{sec:problem} Suppose a user needs to find resources that
provide some functionality: e.g., a service that returns current
weather conditions, or latitude and longitude of a given address. In
order to improve robustness and data coverage of an application, we
often want more than one resource with the necessary functionality.
In this paper, for simplicity, we assume that the user provides an
example resource, that we call a \emph{seed}, and wants to find more
resources with the same functionality. By ``same'' we mean a
resource that will accept the same input data types as the seed, and
will return the same data types as the seed after applying the same
operation to them.
% KL - Should we just define the problem as follows?
Note that we could have a more stringent requirement that the
resource return the same data as the seed for the same input, but
we don't want to exclude resources that may have different
coverage.

We claim that users in a social bookmarking system such as
\emph{del.icio.us} annotate resources according to their
functionality or topic (category). Although \emph{del.icio.us} and
similar systems provide different means for users to annotate
document, such as notes and tags, in this paper we focus on
utilizing the tags only.  Thus, the variables in our model are
resources $R$, users $U$ and tags $T$. A bookmark $i$ of resource
$r$ by user $u$ can be formalized as a tuple $\langle r,u,\{t_{1},
t_{2},\ldots \}\rangle_{i}$, which can be further broken down into
a co-occurrence of a triple of a resource, a user and a tag:
$\langle r,u,t \rangle$.

We collect these triples by crawling \emph{del.icio.us}. The
system provides three types of pages: a tag page --- listing all
resources that are tagged with a particular keyword; a user page
--- listing all resources that have been bookmarked by a
particular user; and a resource page --- listing all the tags the
users have associated with that resource. \emph{del.icio.us} also
provides a method for navigating back and forth between these
pages, allowing us to crawl the site. Given the seed, we get what
\emph{del.icio.us} shows as the most popular tags assigned by the
users to it. Next we collect other resources annotated with these
tags. For each of these we collect the resource-user-tag triples.
We use these data to discover resources with the same
functionality as the seed, as described below.

\section{Approach}
\label{sec:approach} We use probabilistic models in order to find
a compressed description of the collected resources in terms of
topic descriptions. This description is a vector of probabilities
of how a particular resource is likely to be described by
different topics. The topic distribution of the resource is
subsequently used to compute similarity between resources using
Jensen-Shannon divergence \cite{JSDiverge}. For the rest of this
section, we describe the probabilistic models. We first briefly
describe two existing models: the probabilistic Latent Semantic
Analysis (pLSA) model and the Three-Way Aspect model (MWA). We
then introduce a new model that explicitly takes into account
users' interests and resources' topics. We compare performance of
these models on the three \emph{del.icio.us} datasets.

\begin{figure}
\begin{center}
\includegraphics[width=2.6in]{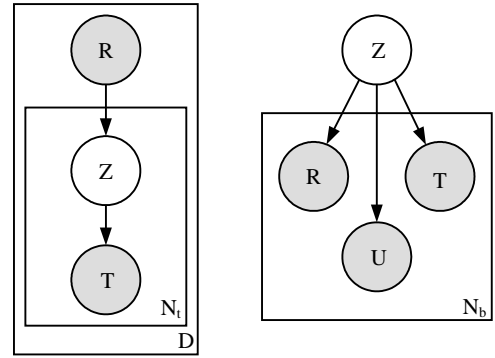}
\end{center}
\caption {Graphical representations of the probabilistic Latent
Semantic Model (left) and Multi-way Aspect Model (right) $R$, $U$,
$T$ and $Z$ denote variables ``Resource'', ``User'', ``Tag'' and
``Topic'' repectively. $N_{t}$ represents a number of tag
occurrences for a particular resource; D represents a number of
resources. Meanwhile, $N_{b}$ represents a number of all
resource-user-tag co-occurrences in the social annotation system.
Note that filled circles represent observed variables.
 } \label{fig:f1}
\end{figure}

\subsection{Probabilistic Latent Semantic Model (pLSA)}
\label{sec:plsa} Hoffman \cite{HofmannPLSA99} proposed a
probabilistic latent semantic model for associating word-document
co-occurrences. The model hypothesized that a particular document is
composed of a set of conceptual themes or topics $Z$. Words in a
document were generated by these topics with some probability. We
adapted the model to the context of social annotation by claiming
that all users have common agreement on annotating a particular
resource. All bookmarks from all users associated with a given
resource were aggregated into a single corpus. \figref{fig:f1} shows
the graphical representation of this model. Co-occurrences of a
particular resource–-tag pair were computed by summing
resource-user-tag triples $\langle r,u,t\rangle$  over all users.
The joint distribution over resource and tag is

\begin{equation}
\label{eq:plsajdist} p(r,t) = \sum_{z} p(t|z)p(z|r)p(r)
\end{equation}

In order to estimate parameters $p(t|z)$, $p(z|r)$, $p(r)$ we define
log likelihood $L$, which measures how the estimated parameters fit
the observed data

\begin{equation}
\label{eq:plsall} L = \sum_{r,t} n(r,t)log(p(r,t))
\end{equation}
\noindent where $n(r,t)$ is a number of resource-tag co-occurrences. The EM
algorithm~\cite{EMpaper} was applied to estimate those parameters
that maximize $L$.

\subsection{Three-way Aspect Model (MWA)}
\label{sec:mwa} The three-way aspect model (or multi-way aspect model, MWA) was originally applied to
document recommendation systems~\cite{Popescul01}, involving 3 entities: user, document and word.
 The model takes into account both user interest (pure collaborative filtering) and document
 content (content-based). Recently, the three-way aspect model was applied on social annotation data in order to
 demonstrate emergent semantics in a social annotation system and to use these semantics for information
 retrieval~\cite{ChineseModel}. In this model, conceptual space was introduced as a latent
 variable, $Z$,
 which independently generated occurrences of resources, users and tags for a particular
triple $\langle r,u,t\rangle$  (see \figref{fig:f1}). The joint
  distribution over resource, user, and tag  was defined
as follows

\begin{equation}
\label{eq:mwadist} p(r,u,t) = \sum_{z} p(r|z)p(u|z)p(t|z)p(z)
\end{equation}

Similar to pLSA, the parameters $p(r|z)$, $p(u|z)$, $p(t|z)$ and
$p(z)$ were estimated by maximizing the log likelihood objective
function, $
%\begin{equation}
%\label{eq:mwall}
L =  \sum_{r,u,t} n(r,u,t)log(p(r,u,t))
%\end{equation}
$. EM algorithm was again applied to estimate those parameters.

\subsection{Interest-Topic Model (ITM)}
\label{sec:dcp} The motivation to implement the model proposed in
this paper comes from the observation that users in a social
annotation system have very broad interests.
% KL - need to develop this - appeal to cognitive studies, mental models, categories
%A particular bookmark
A set of tags in a particular bookmark could reflect both users'
interests and resources' topics. As in the three-way aspect model,
using a single latent variable to represent both ``interests'' and
``topics'' may not be appropriate, as intermixing between these two
may skew the final similarity scores computed from the topic
distribution over resources.

\begin{figure}
\begin{center}
\includegraphics[width=1.6in]{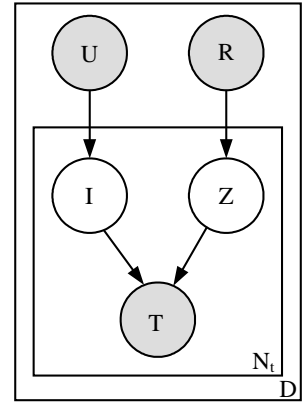}
\end{center}
\caption {Graphical representation on the proposed model. $R$, $U$,
$T$, $I$ and $Z$ denote variables ``Resource'', ``User'', ``Tag'',
``Interest'' and  ``Topic'' repectively. $N_{t}$ represents a number
of tag occurrences for a one bookmark (by a particular user to a
particular resource); D represents a number of all bookmarks in
social annotation system.
 } \label{fig:f2}
\end{figure}

Instead, we propose to explicitly separate the latent variables into
two: one representing user interests, $I$; another representing
resource topics, $Z$. According to the proposed model, the process
of resource-user-tag co-occurrence could be described as a
stochastic process:
\begin{itemize}
\item User $u$ finds a resource $r$ interesting and she would like to
bookmark it

\item User $u$ has her own interest profile ${i}$; meanwhile the resource has a set of topics ${z}$.

\item Tag $t$ is then chosen based on  users's interest and resource's
topic

\end{itemize}

The process is depicted in a graphical form in \figref{fig:f2}. From
the process described above, the joint probability of resource, user
and tag is written as

\begin{equation}
\label{eq:dcpdist} P(r,u,t) = \sum_{i,z}p(t|i,z)p(i|u)p(z|r)p(u)p(r)
\end{equation}

Log likelihood $L$ is used as the objective function to estimate all
parameters. Note that $p(u)$ and $p(r)$ could be obtained directly
from observed data –-- the estimation thus involves three parameters
$p(t|i,z)$, $p(i|u)$ and $p(z|r)$. $L$ is defined as

\begin{equation}
\label{eq:dcpll}
    L = \sum_{r,u,t}n(r,u,t)log(p(r,u,t))
\end{equation}

EM algorithm is applied to estimate these parameters. In the
expectation step, the joint probability of hidden variables $I$ and
$Z$ given all observations is computed as

\begin{equation}
\label{eq:dcpestep} p(i,z| u,r,t) =
\frac{p(t|i,z)p(i|u)p(z|r)}{\sum_{i,z}p(t|i,z)p(i|u)p(z|r)}
\end{equation}

Subsequently, each parameter is re-estimated using $p(i,z| u,r,t)$
we just computed from the E step

\begin{equation}
\label{eq:dcpmstep1}
p(t|i,z) = \frac{\sum_{r,u}n(r,u,t)p(i,z|u,r,t)}{\sum_{r,u,t}n(r,u,t)p(i,z|u,r,t)}
%p(t|i,z) \propto \sum_{u,r}n(u,r,t)p(i,z|u,r,t)
\end{equation}

\begin{equation}
\label{eq:dcpmstep2}
p(i|u) = \frac{\sum_{r,t}n(r,u,t)\sum_{z}p(i,z|u,r,t)}{n(u)}
%p(i|u) \propto \sum_{z,r,t}n(u,r,t)p(i,z|u,r,t)
\end{equation}

\begin{equation}
\label{eq:dcpmstep3}
p(z|r) = \frac{\sum_{u,t}n(r,u,t)\sum_{i}p(i,z|u,r,t)}{n(r)}
%p(z|r) \propto \sum_{i,u,t}n(u,r,t)p(i,z|u,r,t)
\end{equation}

The algorithm iterates between E and M step until the log
likelihood or all parameter values converges.
%The derivation of EM algorithm may be listed at appendix section.

Once all the models are learned, we use the distribution of topics
of a resource $p(z|r)$ to compute similarity between resources and
the seed using Jensen-Shannon divergence.

\section{Empirical Validation}
\label{sec:validation} To evaluate our approach, we collected data
for three seed resources:
\emph{flytecomm}\footnote{http://www.flytecomm.com/cgi-bin/trackflight/}
\emph{geocoder}\footnote{http://geocoder.us} and
\emph{wunderground}\footnote{http://www.wunderground.com/}. The
first resource allows users to track flights given the airline and
flight number or departure and arrival airports; the second
resource returns coordinates of a given address; while, the third
resource supplies weather information for a particular location
(given by zipcode, city and state, or airport). Our goal is to
find other resources that provide flight tracking, geocoding and
weather information. Our approach is to crawl \emph{del.icio.us}
to gather resources possibly related to the seed; apply the
probabilistic models to find the topic distribution of the
resources; then rank all gathered resources based on how similar
their topic distribution is to the seed's topic distribution. The
crawling strategy is defined as follows: for each seed
\begin{itemize}
\item Retrieve the 20 most popular tags that users have applied to
that resource

\item For each of the tags, retrieve other resources that have been
annotated with that tag

\item For each resource, collect all bookmarks that have been
created for it (i.e., resource-user-tag triples)

\end{itemize}
We wrote special-purpose Web page scrapers to extract this
information from \emph{del.icio.us}. In principle, we could continue
to expand the collection of resources by gathering tags and
retrieving more resources that have been tagged with those tags, but
in practice, even after the small traversal we do, we obtain more
than 10 million triples for the \emph{wunderground} seed.

We obtained the datasets for the seeds \emph{flytecomm} and
\emph{geocoder} in May 2006 and for the seed \emph{wunderground}
in January 2007. We reduced the dataset by omitting low (fewer
than ten) and high (more than ten thousand) frequency tags and all
the triples associated with those tags. After this reduction, we
were left with (a) 2,284,308 triples with 3,562 unique resources;
14,297 unique tags; 34,594 unique users for the \emph{flytecomm}
seed; (b) 3,775,832 triples with 5,572 unique resources; 16,887
unique tags and 46,764 unique users for the \emph{geocoder} seed;
(c) 6,327,211 triples with 7,176 unique resources; 77,056 unique
tags and 45,852 unique users for the \emph{wunderground} seed.

Next, we trained all three models on the data: pLSA, MWA and ITM.
We then used the learned topic distributions to compute the
similarity of the resources in each dataset to the seed, and
ranked the resources by similarity. We evaluated the performance
of each model by manually checking the top 100 resources produced
by the model according to the criteria below:
\begin{itemize}
\item \emph{same}: the resource has the same functionality if it provides an input form that takes the same type of
data as the seed and returns the same type of output data: e.g., a
flight tracker takes a flight number and returns flight status

\item \emph{link-to}: the resource contains a link to a page with
the same functionality as the seed (see criteria above). We can
easily automate the step that check the links for the right
functionality.

%\item \emph{in-domain}: the resource meets this criteria if the page provide "related" functions e.g.
%a coordinate conversion resource of the geocoder seed meets this criteria; nevertheless, a booking flight
%or traveling page of flytecomm seed doesn't meet this criteria.
\end{itemize}

Although evaluation is performed manually now, we plan to automate
this process in the future by using the form's metadata to predict
semantic types of inputs~\cite{Hess2003}, automatically query the
source, extract data from it and classify it using the tools
described in \cite{Gazen2005,Lerman2006aaai}. We will then be able
to validate that the resource has functionality similar to the
seed by comparing its input and output data with that of the
seed~\cite{Carman2007}. Note that since each step in the automatic
query and data extraction process has some probability of failure,
we will need to identify many more relevant resources than
required in order to guarantee that we will be able to
automatically verify some of them.

\begin{figure*}[tb]
\begin{tabular}{ccc}
%\epsfxsize = 2in {\epsffile{flytecomm.eps}} & \epsfxsize
%=2in{\epsffile{geocoder.eps}} & \epsfxsize
%=2in{\epsffile{wunderground.eps}}
\includegraphics[width=2.0in]{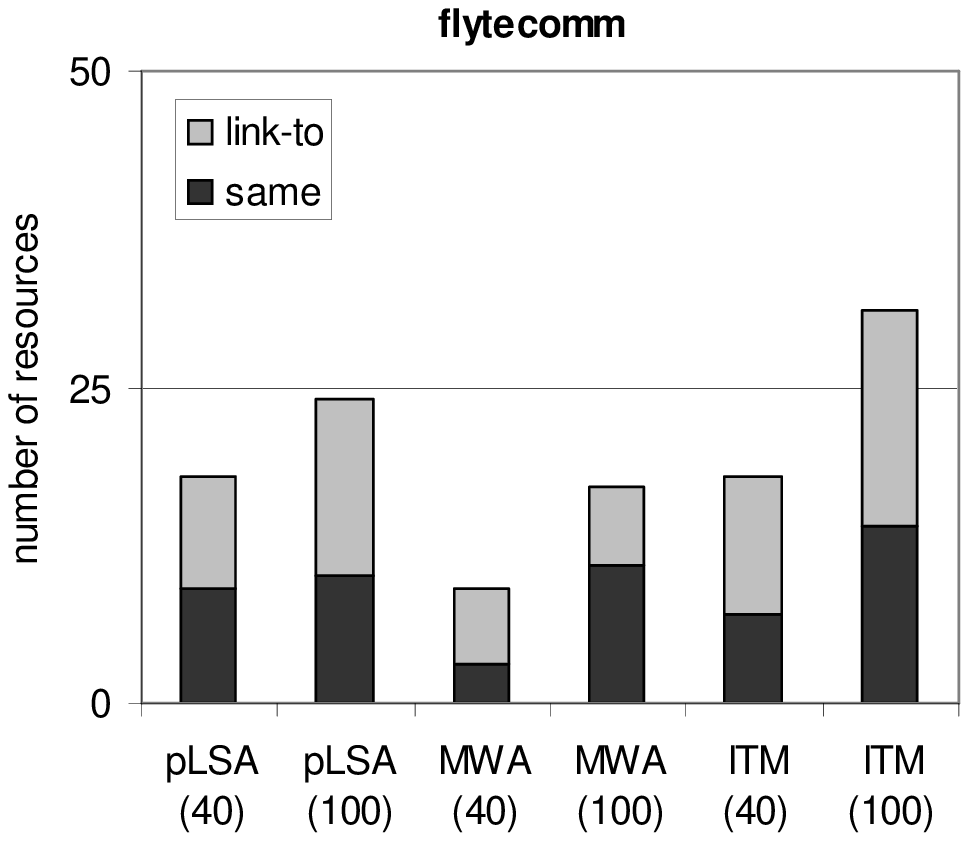} &
\includegraphics[width=2.0in]{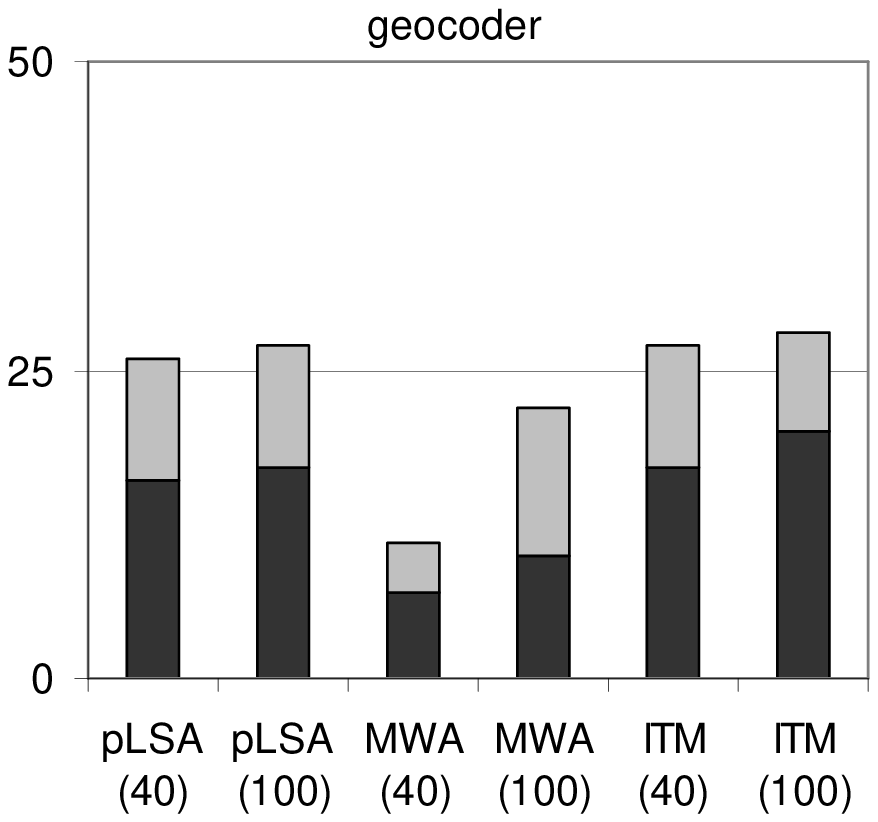} &
\includegraphics[width=2.0in]{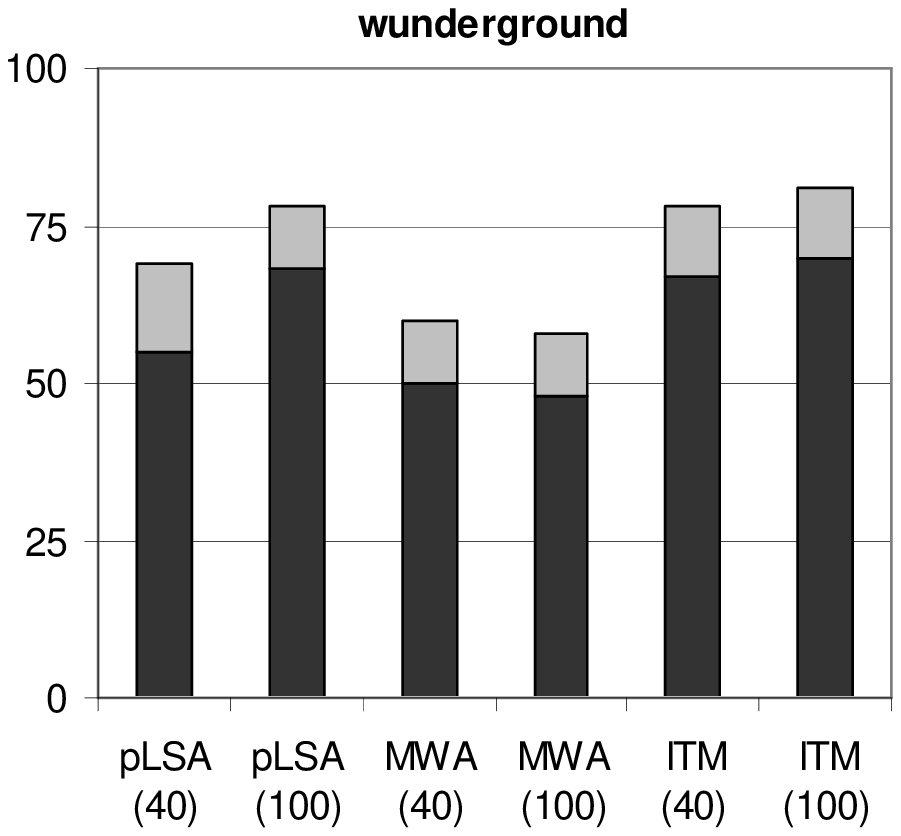}
\end{tabular}
\caption {Performance of different models on the three datasets.
Each model was trained with 40 or 100 topics. For ITM, we fix
interest to 20 interests across all different datasets.  The bars
show the number of resources within the top 100 returned by each
model that had the same functionality as the seed or contained a
link to a resource with the same functionality as the seed. }
\label{fig:f3}
\end{figure*}

Figure~\ref{fig:f3} shows the performance of different models
trained with either 40 or 100 topics (and interests) on the three
datasets. The figure shows the number of resources within the top
100 that had the same functionality as the seed or contained a
link to a resource with the same functionality. The Interest-Topic
model performed slightly better than pLSA, while both ITM and pLSA
significantly outperformed the MWA model. Increasing the
dimensionality of the latent variable $Z$ from 40 to 100 generally
improved the results, although sometimes only slightly. Google's
find ``Similar pages'' functionality returned 28, 29 and 15
resources respectively for the three seeds \emph{flytecomm},
\emph{geocoder} and \emph{wunderground}, out of which 5, 6, and 13
had the same functionality as the seed and 3, 0, 0 had a link to a
resource with the same functionality. The ITM model, in
comparison, returned three to five times as many relevant results.

\begin{table}[b]
\begin{tabular}{|r|r|r|r|r|}
\hline
&PLSA&MWA&ITM&GOOGLE*\\ \hline
flytecomm&23&65&15&$>28$\\
geocoder&14&44&16&$>29$\\
wunderground&10&14&10&10\\
\hline
\end{tabular}
\caption {The number of top predictions that have to be examined
before the system finds ten resources with the desired
functionality (the same or link-to). Each model was trained with
100 topics. For ITM, we fixed the number of interests at 20. *Note
that Google returns only 8 and 6 positive resources out of 28 and
29 retrieved resources for flytecomm and geocoder dataset
respectively. }\label{tbl:effort}
\end{table}

\commentout{ In \emph{wunderground} data, we observed that very
unrelated ranked resources appear intermittently in MWA. At 100
topics, we found 11 ranked resources are about date and time
services. Other unrelated resources are, for example, generic
forum and testing network speed. In contrast, both pLSA and ITM
always ranked ``related'' resources. Although those resources do
not provide an exact function to \emph{wunderground}, their
content is generally about weather.

The situation is similar to that of \emph{flytecomm} and
\emph{geocoder} data. MWA seems to provide a greater number of
unrelated resources than the ones highly ranked by pLSA and ITM.
Note that most of unrelated resources ranked by pLSA and ITM are
about travel and flight booking sites for \emph{flytecomm}, and
map sites for \emph{geocoder}. These unrelated resources are
loosely related to the seed in some manner. Inducing such loosely
related resources might be possibly due to users are likely to
use, in addition to specific tags, more general tags to describe a
particular resource.}

\tabref{tbl:effort} provides another view of performance of
different resource discovery methods. It shows how many of the
method's predictions have to be examined before ten resources with
correct functionality are identified. Since the ITM model ranks
the relevant resources highest, fewer Web sites have to be
examined and verified (either manually or automatically); thus,
ITM is the most efficient model.

One possible reason why ITM performs slightly better than pLSA
might be because in the datasets we collected, there is low
variance of user interest. The resources were gathered starting
from a seed and following related tag links; therefore, we did not
obtain any resources that were annotated with different tags than
the seed, even if they are tagged by the same user who bookmarks
the seed. Hence user-resource co-occurrences are incomplete: they
are limited by a certain tag set. pLSA and ITM would perform
similarly if all users had the same interests. We believe that ITM
would perform significantly better than pLSA when variation of
user interest is high. We plan to gather more complete data to
weigh ITM behavior in more detail.

Although performances  pLSA and ITM are only slightly different,
pLSA is much better than ITM in terms of efficiency since the
former ignores user information and thus reduces iterations
required in its training process. However, for some applications,
such as personalized resource discovery, it may be important to
retain user information. For such applications the ITM model,
which retains this information, may be preferred over pLSA.

\commentout{ We postulate that MWA performs worse than ITM  due to
the former does not explicitly discriminating between user
interests and resource topics. Meanwhile, it performs worse that
pLSA possibly because variation of user interests. In pLSA, we
view all bookmarks generated by a single user or global behavior.
Most consented tags across all users would significantly influence
to a description of a particular resource. Thus common agreement
may dominate variation of user interests, yielding pLSA less
suffering from such variation.}

\commentout{
Other factors that contribute to the model's performance, such as
the number of topics and interests chosen, remain to be
investigated.
}

\section{Previous Research}
\label{sec:previous}

Popular methods for finding documents relevant to a user query rely
on analysis of word occurrences (including metadata) in the document
and across the document collection. Information sources that
generate their contents dynamically in response to a query cannot be
adequately indexed by conventional search engines. Since they have
sparse metadata, the user has to find the correct search terms in
order to get results.

% Hess' work doesn't find similar operations but input/output. Meanwhile, it's a supervised approach
%An early attempt \cite{Hess2003}used metadata contained in the HTML forms of information sources to
%classify the input data types accepted by these sources.

A recent research \cite{Woogle} proposed to utilize metadata in the
Web services' WSDL and UDDI files in order to find Web services
offering similar operations in an unsupervised fashion. The work is
established on a heuristic that similar operations tend to be
described by similar terms in service description, operation name
and input and output names. The method uses clustering techniques
using cohesion and correlation scores (distances) computed from
co-occurrence of observed terms to cluster Web service operations.
In this approach, a given operation can only belong to a single
cluster. Meanwhile, our approach is grounded on a probabilistic
topic model, allowing a particular resource to be generated by
several topics, which is more intuitive and robust. In addition, it
yields a method to determine how the resource similar to others in
certain aspects.

Although our objective is similar, instead of words or metadata
created by the \emph{authors} of online resources, our approach
utilizes the much denser descriptive metadata generated in a social
bookmarking system by the \emph{readers} or \emph{users} of these
resources. One issue to be considered is the metadata cannot be
directly used for categorizing resources since they come from
different user views, interests and writing styles. One needs
algorithms to detect patterns in these data, find hidden topics
which, when known, will help to correctly group similar resources
together. We apply and extend the probabilistic topic model, in particular
pLSA \cite{HofmannPLSA99} to address such issue.

Our model is conceptually motivated by the Author-Topic
model~\cite{TopicModelSmyth2004}, where we can view a user who
annotate a resource as an author who composes a document. The aim in
that approach is to learn topic distribution for a particular
author; while our goal is to learn the topic distribution for a
certain resource. Gibbs sampling was used in parameter estimation
for that model; meanwhile, we use the generic EM algorithm to
estimate parameters, since it is analytically straightforward and
ready to be implemented.

The most relevant work, \cite{ChineseModel}, utilizes multi-way
aspect model on social annotation data in \emph{del.icio.us}. The model
doesn't explicitly separate user interests and resources topics as our
model does. Moreover, the work focuses on emergence of semantic and personalized
resource search, and is evaluated by demonstrating that it can alleviate
a problem of tag sparseness and synonymy in a task of searching for resources
by a tag. In our work, on the other hand, our model is applied to search
for resources similar to a given resource.

There is another line of researches on resource discovery that
exploits social network information of the web graph. Google
\cite{Google} uses visitation rate obtained from resources' connectivity
 to measure their popularity. HITS \cite{HitsKlienberg}
also use web graph to rate relevant resources by measuring their
authority and hub values. Meanwhile, ARC \cite{Chakrabarti98} extends HITS
by including content information of resource hyperlinks to improve
system performance. Although the objective is somewhat similar,
our work instead exploits resource metadata generated by community to
compute resources' relevance score.

% see page 423 of chinesemodel's papar.. their methodology is simple & unrealistic.
% tag they used for testing are "google", "delicious", "java", "p2p", and "mp3" which are very commonly used and over generalized

\section{Conclusion}
\label{sec:conclusion} We have presented a probabilistic model
that models social annotation process and described an approach to
utilize the model in the resource discovery task. Although we
cannot compare to performance to state-of-the-art search engine
directly, the experimental results show the method to be
promising.

There remain many issues to pursue. First, we would like to study
the output of the models, in particular, what the user interests
tell us. We would also like to automate the source modeling process
by identifying the resource's HTML form and extracting its metadata.
We will then use techniques described in \cite{Hess2003} to predict
the semantic types of the resource's input parameters. This will
enable us to automatically query the resource and classify the
returned data using tools described in
\cite{Gazen2005,Lerman2006aaai}. We will then be able to validate
that the resource has the same functionality as the seed by
comparing its input and output data with that of the
seed~\cite{Carman2007}. This will allow agents to fully exploit our
system for integrating information across different resources
without human intervention.

\commentout{ Our next goal is not only to find resources which have
similar functions, but also to find composable resources. We do need
our system to be able to find related resources in different
resolutions.
 From datasets we collected, although there is no explicit hierarchical
 order for tags, users in social annotation system generally provide
 a set of tags composing of both generic and specific ones. For example
 several users use ``travel'' tag and ``flighttracking'' tag together
 to describe resources providing online flight tracking data. Although
 travel is not specifically relevant to flight tracking but ``travel''
 may be used by other users who describe other resources that are
 composable with the flight tracking data. Probabilistic topic model
 which takes care of topic correlations such as Pachinko Allocation,
\cite{Pachinko06}, is thus a promising to be applied. }

Our next goal is to generalize the resource discovery process so
that instead of starting with a seed, a user can start with a query
or some description of the information need. We will investigate
different methods for translating the query into tags that can be
used to harvest data from \emph{del.icio.us}. In addition, there is
other evidence potentially useful for resource categorization such
as user comments, content and input fields in the resource. We plan
to extend the present work to unify evidence both from annotation
and resources' content to improve the accuracy of resource
discovery.

\paragraph{Acknowledgements}
This research is based by work supported in part by the
NSF under Award No. CNS-0615412 and in part by DARPA under Contract No. NBCHD030010.

\bibliographystyle{aaai}
\bibliography{delicious}

\end{document}